# USING AI TO REPLICATE HUMAN EXPERIMENTAL RESULTS: A MOTION STUDY

Rosa Illán Castillo[1]

Javier Valenzuela[2]

[1] CNRS – Laboratoire Dynamique Du Langage (UMR 5596)
maria-del-rosario.illan-castillo@cnrs.fr

[2] Universidad de Murcia
jvalen@um.es

**Abstract:** This paper explores the potential of large language models (LLMs) as reliable analytical tools in linguistic research, focusing on the emergence of affective meanings in temporal expressions involving manner-of-motion verbs. While LLMs like GPT-4 have shown promise across a range of tasks, their ability to replicate nuanced human judgments remains under scrutiny. We conducted four psycholinguistic studies—on emergent meanings, valence shifts, verb choice in emotional contexts, and sentence-emoji associations—first with human participants and then replicated the same tasks using an LLM. Results across all studies show a striking convergence between human and AI responses, with statistical analyses (e.g., Spearman's $\rho$ = .73–.96) indicating strong correlations in both rating patterns and categorical choices. While minor divergences were observed in some cases, these did not alter the overall interpretive outcomes. These findings offer compelling evidence that LLMs can augment traditional human-based experimentation, enabling broader-scale studies without compromising interpretive validity. This convergence not only strengthens the empirical foundation of prior human-based findings but also opens possibilities for hypothesis generation and data expansion through AI. Ultimately, our study supports the use of LLMs as credible and informative collaborators in linguistic inquiry.

## 1 The role of AI in linguistic research

There are clear bottlenecks in the acquisition of linguistic data: our current corpora provide linguists with a staggering amount of results, which then have to be painstakingly analyzed manually for the most part (Crosthwaite 2023). In psycholinguistic experimentation, the number of stimuli which can be studied is limited by factors related to human performance, such as attention or concentration (Trott 2024a). The emergence of AI, particularly large language models (LLMs) such as GPT-4, Gemini or Llama, with their stunning capabilities, opens up new possibilities to circumvent such bottlenecks (e.g., Aher et al. 2022; Alzahrani 2025; Heyman and Heyman 2024; Torrent et al. 2023).

However, despite this potential, the use of AI as an analytical tool in linguistics has not yet become widespread. The main reason probably concerns the reliability of the performance of LLMs: if LLMs provide judgments that are sufficiently humanlike, then the current datasets could be safely expanded, with more words and more analytical dimensions. Should LLMs provide responses too different from those of humans, the whole enterprise would have to be put on hold until their performance matches humans to a level considered acceptable.

In this paper, we present the results of a comparative study of the data obtained from four human-based experiments and the output generated by AI for the same experimental tasks. The experiments were centered on a very concrete topic: the emergence of affective meanings when manner of motion verbs are used in temporal expressions. In time conceptualization, details about the path of motion are easily mapped (the source/goal of the motion becomes the beginning/end of the temporal stretch; the space in front/behind of the speaker becomes the future/past; "here" becomes "now", etc.). However, mapping manner of motion information is not as immediately straightforward. Previous studies have observed that specific details of manner of motion are typically mapped onto our subjective perception of time (Authors 2022): e.g., whether time is conceptualized as going fast or slow (e.g. *time flies* vs. *minutes drag*). There are also a number of emergent meanings which are triggered by these verbs: verbs like *drag* can evoke a sense of boredom or anguish; verbs like *fly* are used to indicate that you are having fun, or signal a very active period; verbs like *slip* are associated with the notion that the passage of time has gone undetected, escaping the conceptualizer's attention. The four experiments were aimed at testing nuanced evaluations of concrete, contextualized examples with these verbs, using different tasks, such as elicitation of affective meanings, an exploration of the modulation of the valence of these verbs in metaphorical contexts and association with different emojis. The question that is asked in this paper is whether the same conclusions would be reached if we used the results provided by the LLM instead of those from human participants.

## 2 Experimental studies and their replication with AI

### 2.1 Study 1. Emergent meanings survey

#### 2.1.1 Human-based study

*Rationale*
This study was designed to explore how different features associated with manner-of-motion verbs, particularly speed, contribute to meaning construction and trigger different emotional interpretations in temporal contexts.

*Participants*
The study involved 59 native English speakers, selected from the platform Prolific. 39 were female and 20 were male; their ages ranged between 22 and 54.

*Stimuli*
There were 15 sentences, each featuring a motion verb with a Time Measurement Unit (TMU) as their subject (see Appendix 1). TMUs are linguistic expressions that quantify time, such as *minutes*, *hours*, *days*, and *years*. The verbs selected for this study included *approach*, *fly*, *hasten*, *race*, *rush*, *slip*, *slide*, *spin*, *drag*, *creep*, *crawl*, *limp*, *edge [closer]*, *inch [closer]*, and *run*.

*Procedure*

The survey was administered online via Qualtrics. Participants were instructed to read the 15 sentences carefully and then respond to a series of 9 questions regarding the temporal and emotional implications conveyed by the use of the motion verb in context (see Appendix 1). For each question, a slider ranging from 1 to 7 allowed participants to quantify their perceptions, with an additional option to select *Not apply* where appropriate. The questions aimed to capture a spectrum of temporal experiences, from the speed of time's passage to the emotional and situational context implied by the verb's use.

*Results*

To start with, the results showed that verbs indicative of faster motion (e.g., *fly*, *hasten*, *race*, *rush*, *slip*, *slide*, *spin*) were associated with subjective perceptions of quicker time passage, whereas verbs indicating slower movement (e.g., *drag*, *creep*, *crawl*, *limp*) correlated with slower time perceptions. Speed was thus identified as the central manner feature that activates further emotional meanings related to the speaker's interpretation of time. The responses also showed that sentences with verbs implying a *fast pace* were rated as more lively and entertaining and were associated with narratives that were perceived as more eventful (more things happening); they were also negatively correlated with perceived difficulty (that is, faster motion verbs were associated to a more streamlined progression of events). Additionally, faster verbs induced a higher sense of agency, of "control" over events happening.

In summary, the results of Study 1 suggested that speed is more than a descriptive attribute shaping the emotional quality of temporal experiences and the perceived control over those experiences. Faster verbs suggest greater agency, eventfulness, and anticipation, while slower verbs convey monotony and a more passive experience of time.

### 2.1.2 LLM test

*System*

After some trials we settled on ChatGPT o1 for this task.

*Prompt*

The prompt presented to the system was exactly the one presented to human subjects.

*Results*

The results converged to a great degree with the ones provided by humans. If we look at the differences per verb (averaging all the questions), the results are very similar:

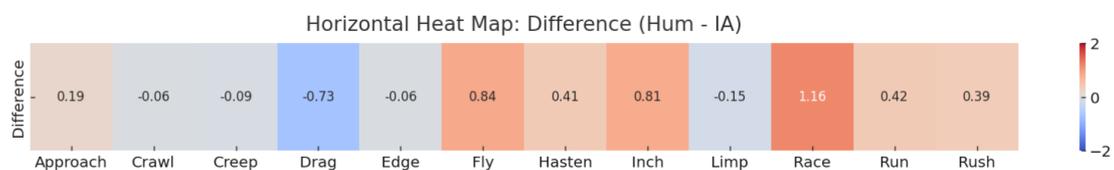

Figure 1. Heatmap with differences in responses of human participants and AI system averaged by verb

The values ranged from 0.05 (*edge*) and 0.06 (*crawl*), to the difference corresponding to the verb *race* (1.16). This difference shows up again when we take a closer look at the responses of the different verbs and different questions (see Figure 2):

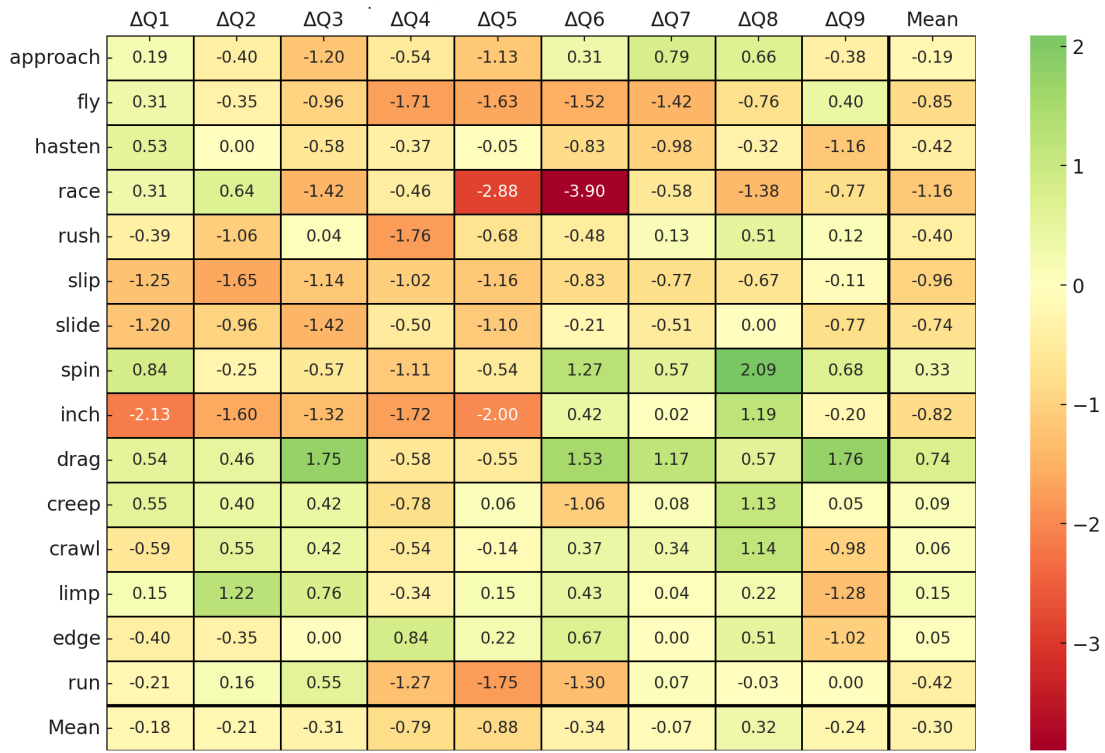

Figure 2. Heatmap with differences of the responses of questions/verbs by human participants and AI system

As it can be seen, most of the deviations stayed below 1 (on a scale of 1 to 7). A great majority of answers fell within the central area of a distributional analysis (approximately 71.11% of the ΔQ values fall within one standard deviation of the mean and 96.30% between two standard deviations). A statistical analysis revealed these differences to be non-significant. The questions with higher divergence corresponded to the verb *race* again; questions 5 and 6 were about the productivity of the speaker during the time period mentioned and the sense of imminence (something is about to happen).

The plot in Figure 3 shows the rank-order relationship between GPT-4's ratings and human ratings across verb-dimension pairs. Each point represents a verb-dimension pair, comparing human and GPT-4 ratings. The Spearman's $\rho = 0.73$ indicates a strong and statistically significant positive correlation, meaning that GPT-4 preserves the relative ordering of human judgments across these items quite well.

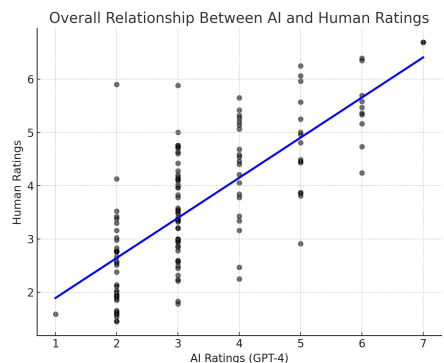

Figure 3. Spearman's correlation of questionnaire responses of human participants and AI system

## 2.2 Study 2: Metaphoric effects on valence

### 2.2.1 Human-based study

*Rationale*
This study examined the emotional valence of manner-of-motion verbs in both literal (physical) and metaphorical (temporal) contexts.

*Participants*
Forty-six native English speakers, selected from the platform Prolific; 23 were female and 23 male, and their ages ranged between 21 to 45.

*Stimuli*
There were 20 sentences which included 10 manner of motion verbs (*crawl, creep, drag, edge, fly, inch, limp, linger, blow* and *zoom*), used in both literal and metaphorical conditions (see Appendix 2). Each verb appeared twice (once for each condition). The physical and metaphorical versions of the sentences contained the same number of words and their structures were kept as similar as possible (e.g., *the birds flew by* vs *the months flew by*).

*Procedure*
The study was conducted online using Qualtrics. Participants read each sentence carefully and rated its emotional tone using an interactive slider corresponding to the 1-to-9 valence scale. The instructions emphasized that there were no right or wrong answers and that participants should base their ratings solely on their personal interpretation of each sentence. To prevent repetition effects, each participant viewed only one of the two conditions (literal or temporal) for each verb. Sentences were presented in random order to control for order effects.

*Experimental results*
Our results showed that participants did not rate the same verbs in the same way in either physical or metaphorical contexts. Instead, there appeared a very consistent variation by which "fast verbs" (e.g., *fly, zoom, blow*) increased their valence ratings in metaphorical settings (that is, were seen as more positive), while "slow verbs" (e.g., *drag, creep, inch*) were rated as more negative in metaphorical settings.

### 2.2.2 LLM test

*System.*
For this test, we used again ChatGPT o1

*Prompt*
The prompt presented to the system was exactly the one presented to human subjects

*Results*
As in the case of human participants, the results provided by the AI system showed that the associated speed of the verb was the factor that tilted the valence in a metaphorical setting. Though the ratings were not exactly the same (for example, slow verbs in physical settings received a rating of 4.56 by humans and a 3.14 by the AI system), the pattern in valence variation was the same in both cases. The ratings were decreased in metaphorical settings by an almost equivalent account: humans lowered the rating by 0.95 and the AI system did the same by 0.86 (see Figure 4). The same tendency was observed in the case of quick verbs; though the initial ratings in physical settings were slightly different (5.17 humans vs 5.42 AI), these ratings were increased in metaphorical settings by both humans and AI.

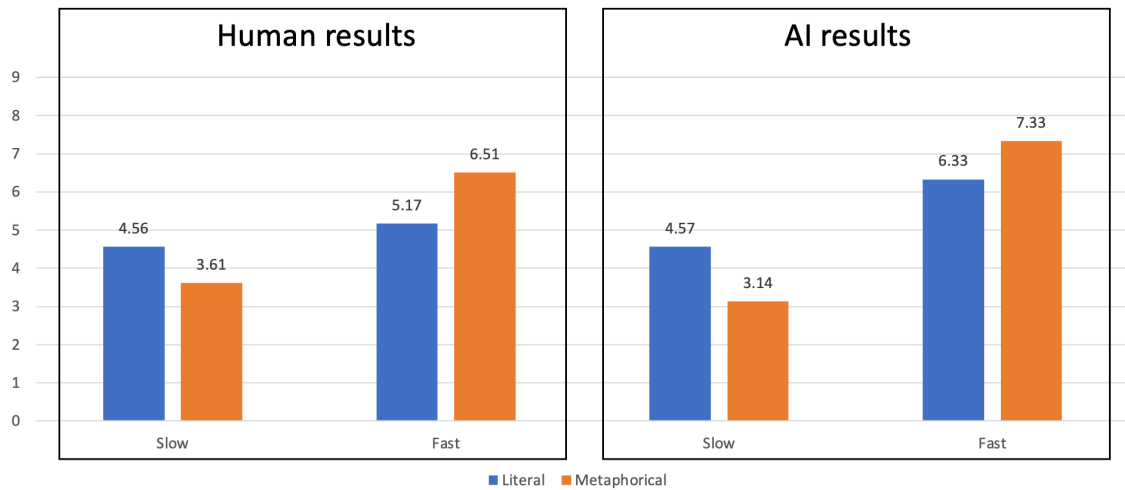

Figure 4. Valence ratings of slow and quick verbs in literal and metaphorical contexts by human participants and the AI system

We submitted the results to a Spearman's correlation test and the ρ value (0.96) indicated an extremely strong positive rank-order correlation between GPT-4 and human ratings (see Figure 5). This plot shows that GPT-4's ratings preserve the relative ranking of human judgments with remarkable accuracy. This very high Spearman's ρ suggests that GPT-4 is not just broadly aligned with human interpretation but can also accurately reflect subtle distinctions in perceived intensity or nuance across items.

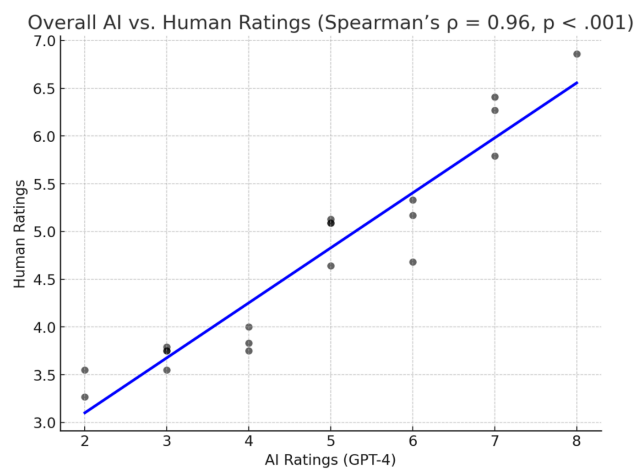

Figure 5. Spearman correlation of valence responses of human participants and AI system

### 2.3 Study 3: Fill in the blanks with Path and Manner verbs

#### 2.3.1 Human-based study

*Rationale*

This study used a fill-in-the-blank task to examine participants' preferences for manner versus path verbs in sentences with explicitly encoded emotions.

*Participants*
The study involved 32 native English speakers aged between 18 and 27; 24 were females and 9 were males. All participants were undergraduate students in the Department of Cognitive Science at the University of California, San Diego

*Stimuli*
The material consisted of 12 sentences with a missing verb. The sentences mentioned explicitly some emotion (e.g. *His speech was monotonous. Those two hours_____(dragged/passed) by* or *We had a wonderful time at the beach. It felt like the hours _____ (crawled/rushed) by*. Participants had to choose one verb out of a pool of manner and path verb (see Appendix 3).

*Procedure*
Participants were presented with the 12 sentences on a computer screen, through Qualtrics, one at a time. Each had a blank space and two options; participants had to select the motion verb that best suited the context of the sentence. Half of the sentences included a choice between two manner verbs, and the other half offered a choice between a manner and a path verb.

*Experimental results*
Results showed a clear congruence between the verbs chosen and the emotional content of the sentence. When choosing between two manner verbs, their selections consistently aligned with the emotional content expressed in the sentences. For example, for a sentence depicting a tedious event, *limping* was widely preferred over *flying*. For enjoyable events, verbs such as *rush* and *zoom* were favored. When choosing between a manner verb and a path verb, participants invariably opted for the manner verb. This suggests a preference for manner verbs when emotional content is present, highlighting the cognitive inclination to emphasize the emotional aspects of an experience when describing the passage of time.

### 2.3.2 LLM test

*System*
ChatGPT o1 was used for this task.

*Prompt*
The prompt presented to the system was exactly the one presented to human subjects

*Results*
The options judged as more natural by the AI system coincided up to a 100% with the options chosen by the human participants (see Figure 6).

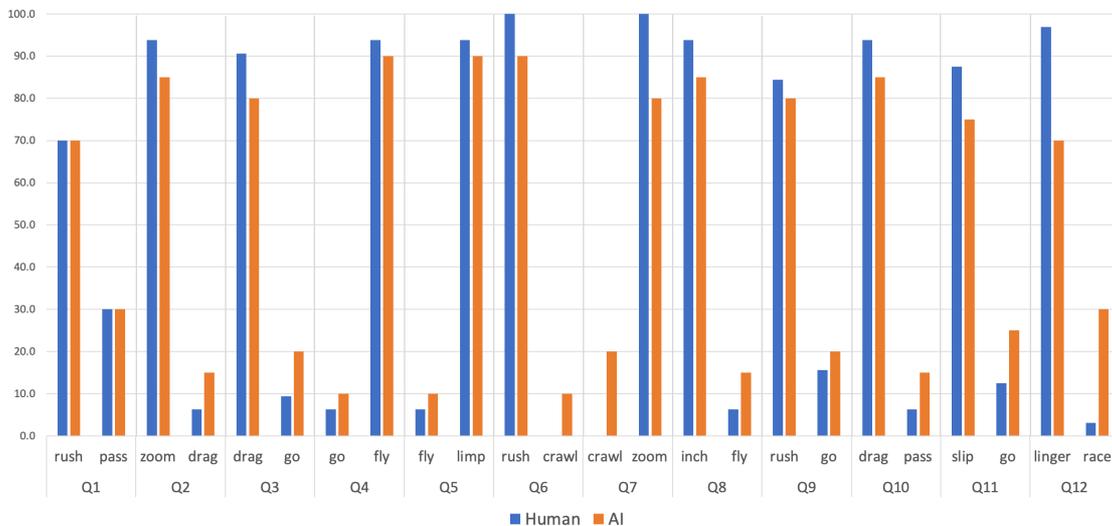

Figure 6. Choices of humans and AI system in the forced choice fill-in-the-blanks task

Both the humans and the AI made always the same choices; that is, the verb they preferred was always the same one. In the case of humans, there were some differences in the strength of the chosen verb (that is, not all the options were chosen by 100% of participants). In order to mimic this result in ChatGPT, we asked the system to rate the probability of the option chosen and then we compared that with the number of participants that had chosen one of the options. Generally speaking, the differences were minor. In one case, the values were exactly the same: 70% of the participants chose *rush* as their preferred verb for Q1, while the other 30% went for *pass*, and ChatGPT provided exactly the same numbers. In 8 out of the 10 cases, the differences were small, with a 10-12% maximum variation range (see Figure 7)

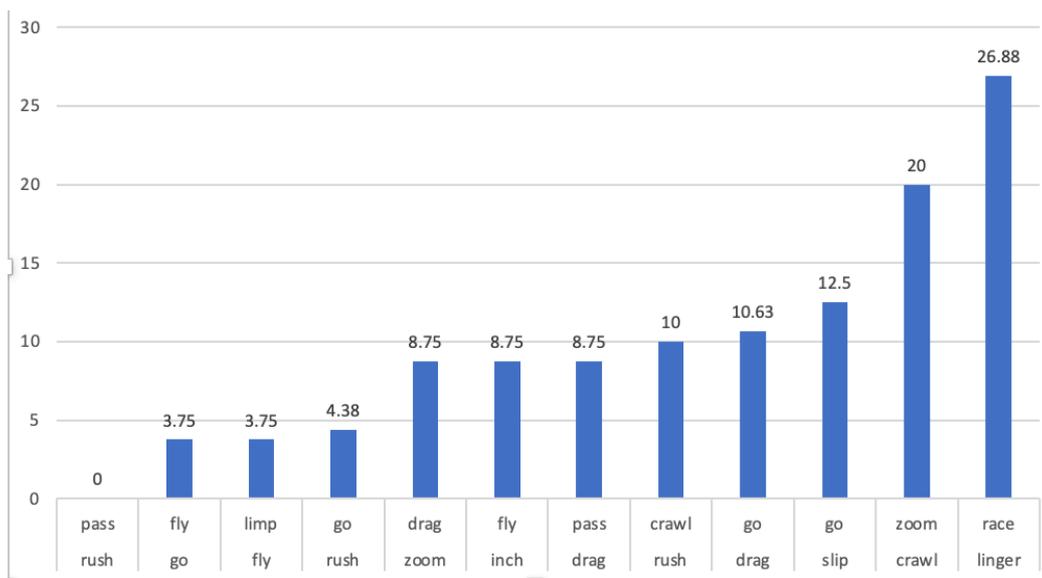

Figure 7. Difference between probability choices of humans and AI

In the two remaining cases, humans went for their preferred option much more clearly than the AI system. In the case of *crawl-zoom*, 100% of humans chose *zoom* while ChatGPT gave it a 80% possibility, a difference of 20 points. This was even more accused for the pair *linger-race*; 96,9% of humans chose *linger*, while ChatGPT gave it a 70% possibility (26.88% difference).

## 2.4 Study 4. Sentence-emoji association

### 2.4.1 Human-based study

*Rationale*
This study investigated the relationship between metaphorical language and non-verbal emotional representations, with the use of emojis.

*Participants*
This study involved 32 native English speakers aged between 18 and 27; 24 were females and 9 were males. All participants were undergraduate students in the Department of Cognitive Science at the University of California, San Diego

*Stimuli*
The materials for this experiment consisted of 14 sentences and 11 emojis (see Appendix 4. Each sentence incorporated a motion verb used with a TMU serving as the subject. The motion verbs included were *slip, creep, inch, blow, limp, drag, crawl, linger, fly*, and *rush*. The selection of emojis for this experiment was based on a prior validation process to ensure that each emoji's emotional significance was unmistakably aligned with either positive or negative valence. During the validation phase, participants evaluated whether each emoji conveyed positive, negative, or neutral sentiments

*Procedure*
The study was conducted using the Qualtrics platform After reading each sentence, participants were shown two emojis: one congruent and one incongruent with the verb's emotional tone. Participants were tasked with selecting the emoji that best captured the emotion evoked by the preceding sentence.

*Experimental results.*
Results showed a clear link between manner-of-motion verbs and either positive or negative valenced emojis when used in temporal contexts. Verbs such as *drag*, *linger*, *crawl*, *creep*, *limp*, and *inch* (slow verbs) were associated with negative-valenced emojis and verbs such as *fly*, *blow*, and *rush* (quick verbs) were associated with positive-valenced emojis. The verb *slip* was found to have a dual sentiment; this ambiguity was attributed to the nuanced implications *slip* carries, which can convey a sense of a quick passage of time or a loss of control.

### 2.4.2 LLM test

*System.*
ChatGPT o1 was used for this task.

*Prompt*
The prompt presented to the system was exactly the one presented to human subjects.

*Results*
In this case, ChatGPT chose the same emoji humans had chosen 100% of the times. As we did in the previous study, we transformed the choice of humans to a probability calculation by ChatGPT. Humans were somewhat more categorical in their choices; the

AI system, though finally choosing the same emoji, tended to give a higher probability to the alternative choice. Figure 8 summarizes the results

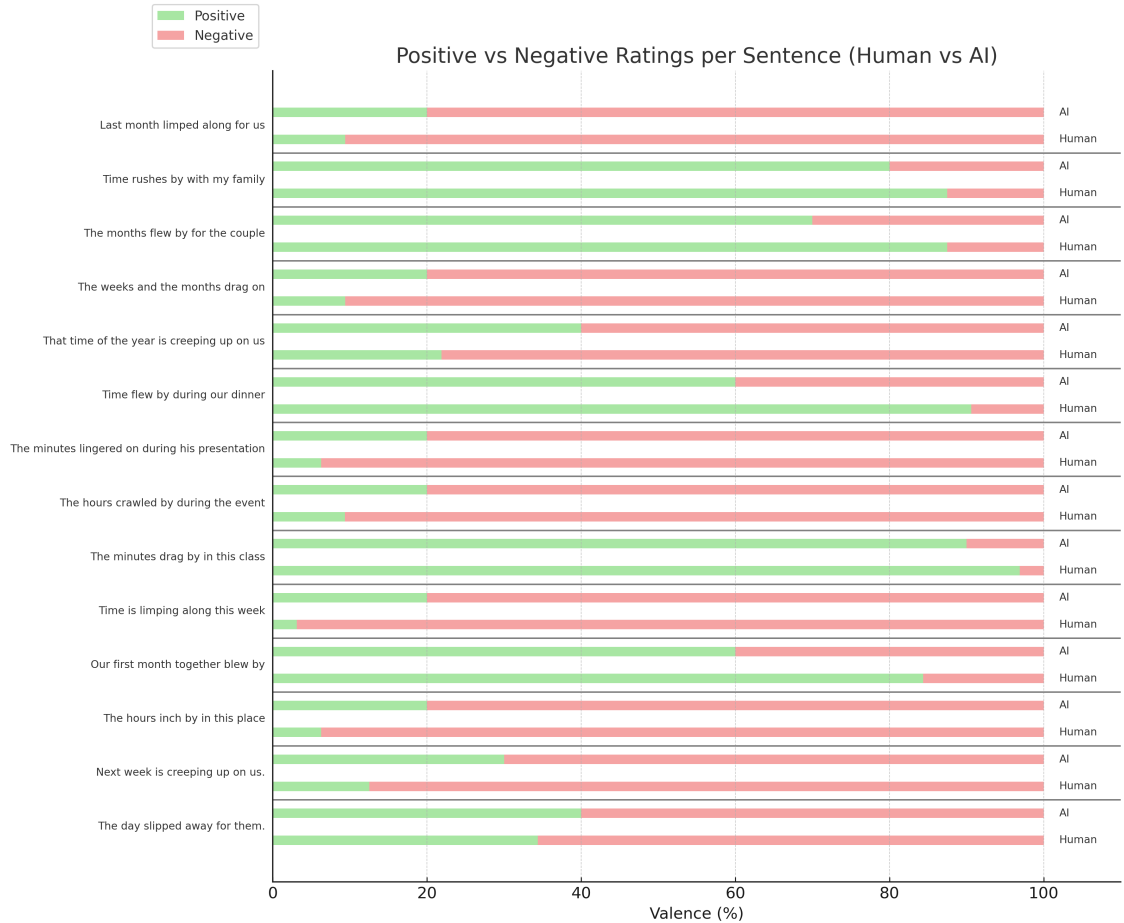

Figure 8. Positive and negative emoji choice by humans and AI

As in the previous studies, we submitted the results to a Spearman's correlation analysis, which reached 0.69 (see Figure 9).

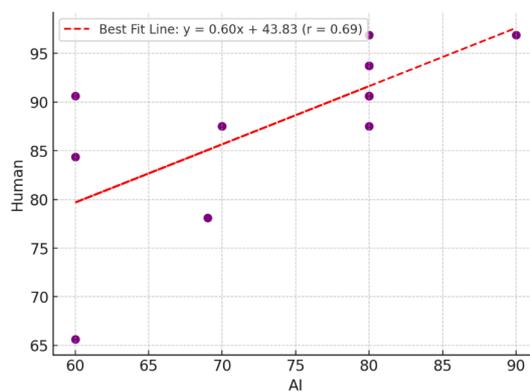

Figure 8. Spearman correlation of emoji choices of human participants and AI system

In this study, we asked the AI system to justify its answers; the system accordingly provided a brief description of each of the emojis in its justification. In this way, we were able to detect that the system had difficulties understanding two of the emojis; when leaving out the values of the emojis the system did not fully understand, the correlation values are increased (y=0.46x + 54.94; r = 0,73).

# 3 Discussion

The results of humans and of the AI system can be regarded as fully compatible. That is, if the study had been conducted only with AI instead of human participants, the conclusions reached would have been the same. Not all tasks proposed to the system were equally challenging for the AI. Clearly, Task 3, the "fill-in-the-blanks" task, was expected to yield positive results, since it mirrors closely what LLMs do. It is thus not surprising that the results match exactly those of human participants. Both Study 1 and Study 2, however, depended on subjective intuitions of the participants. Study 1, for example, required participants to evaluate their emotional and affective reactions to the sentences selected. In this case, it is surprising how well the AI system did, capturing a number of emotional nuances (most of them not found in the dictionary definitions of the verbs), that approached very closely those of human participants. The same goes for Study 2, which asked about valence appraisals of different sentences: here, the polarization reported by humans (increasing the valence of quick verbs in metaphorical settings and decreasing that of slow verbs) was paralleled by the AI system. Probably, this is one of the most important results of the experimental part, since this valence variation could in principle be considered a truly emergent phenomenon; thus it is especially interesting that the human results are fully validated by the AI system. Finally, Study 4, the emoji task, is probably the most challenging for an AI, if only because it requires multimodal abilities: the system has to identify correctly the nature of the emoji before it can relate it to a specific sentence. In this case, the results provided by the AI system were again completely identical those of human participants.

    This is not to say that the results were absolutely identical. In Study 1, the highest divergences were found with the verb *race*, when asking about the "productivity" of the conceptualizer, or the sense of "imminence" conveyed. In Study 3, the choices made by the human participants were generally speaking more categorical than the probabilities provided by the AI system; again, as in Study 1 the most pronounced case corresponded to the verb *race*. Though these differences are not relevant and the study carried out exclusively with the AI system would lead to the same conclusions, it is interesting to consider the nature of these divergences. The fact that verb *race* appears as the most divergent example in two studies is probably not random. About the reason for this, we can only speculate, since these systems are "black boxes" and it is almost impossible at this moment to know exactly which are the paths that guide them to their results. *Race* is a conversion-word which serves as both verb and noun, and quite probably the verb version is less frequent that its noun function, but the same could be said of *inch*, which shows no problems. Since LLMs keep in their vectors very detailed information about the contexts in which words occur, it might be the case that there is information in the noun version of *race* which is tilting the results, again, this is offered as mere speculation. At this moment, the rise of "probing methods" (Belinkov 2022; Zhao et al. 2024), which allow to check whether a particular type of information is present in the vectorial representation of a word, could settle the matter, but these methods are still immature and in an early stage of their development. Regarding the more categorical judgments of humans vs AI, it could be due to the type of prompt used with the system; while the AI system initially provided results which agreed to a 100% in the case of Studies 3 (fill in the blanks) and 4 (emoji association), we attempted to mimic the percentage of choices by humans by asking the AI system to give probabilities to each choice, which could be taken as a rather forced and artificial task and the system could well feel compelled to give other choices some force because of our prompt.

## 4 Conclusion

There are several conclusions we can extract. The first one concerns our central goal. Apparently, and for the tasks we have used to advance our knowledge about the use of manner of motion verbs in temporal contexts and their role in the triggering of emotional effects, the LLM tested has led us to very similar conclusions as the experiments with human subjects. In fact, aside for some small nuances, our results would have been the same if we had used the LLM instead of the human pool. This, by itself, provides our studies with a more solid empirical foundation, since it could be taken as a case of "convergent evidence" (Munafò and Smith 2018) and joins other works (e.g., Gilardi et al. 2023; Törnberg 2023; Trott 2024b) that have argued in this sense. These results open up a thrilling possibility: what if instead of using 15 sentences as stimuli, we used 200? There are only two logical possibilities: either the results would be the same, in whose case, they would increase the empirical base of the explanation and thus the solidity of the conclusions extracted: we would know that these mechanisms work not only with fifteen cases, but with many more. The second possibility is even more intriguing and exciting: we could get new emotional triggerings, new mechanisms suggested by LLMs system, which could then be examined carefully by the scholars and, if accepted as valid, be incorporated into the explanation of the phenomenon, and in turn trigger new experiments and new analyses.

The second conclusion is somehow more theoretical: the fact that results we get using LLMs converge with human judgments is further evidence of the staggering amount of information which is contained in language. A number of scholars (for example, Louwerse 2018, 2021) have been able to replicate many of the experiments cited as evidence for embodiment using only language-based models. In this sense, the amount of information contained in language continues to amaze the scientific community. Probably there is a yet-to-be-discovered amount of knowledge that can be gleaned out of linguistic material. Unfortunately, this can be considered as a classic chicken-and-egg problem: it is not clear whether the information comes from the statistical connections between words themselves—without the need for any embodied simulation, as scholars in more structuralist or symbolists positions would argue, or this information comes from our interaction with the world, and language acts as "a mirror of reality" and that is why this information is here and linguistic patterns just elaborate on this foundational knowledge. The issue of the exact nature of the meanings extracted from LLMs is a thorny issue, which we do not intend to tackle here. What is clear, however, is that this information is in language, regardless of its source, and that our findings support the view that large language models constitute a resource of legitimate interest to linguists.

APPENDIX 1

**Stimuli for study 1: Emergent meanings survey (English)**

*Sentences:*

1. - The new year is approaching.
2. - The two hours that the show lasted flew by.
3. - The minutes hasten in this place.
4. - Our dinner together raced by.
5. - The days rush when I'm home.
6. - Last week slipped by.
7. - The days slid into January.
8. - We will know more as the day spins on.
9. - Election day is inching closer.
10. - The first full day of trial dragged on.
11. - The days creep towards the end of the week.
12. - The hours crawled by that afternoon.
13. - This month is limping along.
14. - The hour edged closer.
15. - The news hour runs from 5pm to 7pm.

*Questions:*

1) How is the passage of time perceived? (Very slowly to Very quickly)
2) Time is passing in a (Monotonous, boring way to Lively, entertaining way)
3) How many things are happening? (Nothing is happening to Lots of events are happening)
4) Are events under control? (No control over the events to Events are fully controlled)
5) How productive is the speaker? (They achieve nothing to They get a lot of things done)
6) Is there a sense of imminence, something is about to happen? (No to Yes)
7) Is there a sense of surprise of the event happening? (No to Yes)
8) Is the speaker impatient of what is to come? (No to Yes)
9) Is time passing with difficulty, are there some setbacks during the day? (No to Yes)

APPENDIX 2

**Stimuli for study 2: Constructing meaning and valence - Manner of motion verbs in literal (physical) and metaphorical (temporal) contexts**

1. **Crawl**
    - Literal: The ant crawled by.
    - Temporal: The hours crawled by.
2. **Creep**
    - Literal: The ivy crept up.
    - Temporal: The minutes crept by.
3. **Drag**
    - Literal: He dragged it along.
    - Temporal: The days dragged along.
4. **Edge**
    - Literal: The turtle edged along.
    - Temporal: The weeks edged along.
5. **Fly**
    - Literal: The plane flew by.
    - Temporal: The years flew by.
6. **Inch**
    - Literal: The snail inched by.
    - Temporal: The hours inched by.
7. **Limp**
    - Literal: He limped forward.
    - Temporal: The days limped forward.
8. **Linger**
    - Literal: His finger lingered on the glass.
    - Temporal: The minutes lingered on.
9. **Zoom**
    - Literal: The motorcycle zoomed by.
    - Temporal: The minutes zoomed by.
10. **Blow**
    - The wind blew by
    - The hours blew by

APPENDIX 3

**Stimuli for study 4: Fill in the blanks with path and manner verbs**

- Exercising with a friend can make the minutes ______ by. (Zoom / Drag)
- Our work last week was tedious. It felt like the days were ______. (Flying by / Limping along)
- We were thrilled with their visit. During our time together the hours ______ by. (Rushed / Crawled)
- The summer camp was fascinating. Three weeks ______ by. (Crawled / Zoomed)
- This place is dull. The hours ______ by. (Inch / Fly)
- For next week's class we must write about a wearisome movie. When I tried to watch it the hours ______. (Lingered on / Raced by)
- While she waited for the police to give her some information about her missing brother, the minutes ______ by. (Dragged / Went)
- Our tour around Rome was captivating. The three days we spent there ______ by. (Rushed / Went)
- It's been just two days and I already love this teacher. During her classes time ______ by. (Goes / Flies).
- His speech was monotonous. Those two hours ______ by. (Dragged / Passed)
- We were not very productive yesterday. Our day ______. (Slipped away / Passed by by)
- We had a wonderful time at the beach. The day ______. (Rushed by / Passed by).

APPENDIX 4

**Stimuli for study 3: Sentence-emoji association**

1) **Sentences:**

- The day slipped away for them.
- Next week is creeping up on us.
- That time of the year is creeping up on us.
- The hours inch by in this place.
- Our first month together blew by.
- Time is limping along this week.
- The minutes drag by in this class.
- The hours crawled by during the event.
- The minutes lingered on during his presentation.
- Time flew by during our dinner.
- The weeks and the months drag on.
- The months flew by for the couple.
- Time rushes by with my family.
- Last month limped along for us.

2) **Emojis:**

- Positive valence:

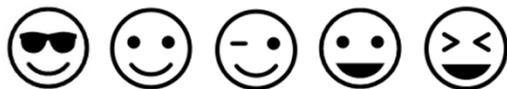

- Negative valence:

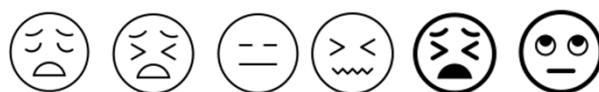